%% file: main.tex
\documentclass[runningheads]{llncs}
\usepackage[T1]{fontenc}
\usepackage{graphicx}
\usepackage{amsmath}
\usepackage{tikz}
\usepackage{pgfplots}
\usepackage{cleveref} 
\newtheorem{asu}{Assumption}
\usepackage{todonotes}
\usepackage{multirow}
\usepackage{url}
\usepackage{amsfonts}
\usepackage{caption}

\begin{document}
\title{Nearest-Neighbor Density Estimation for Dependency Suppression}
%
%
\author{Kathleen Anderson\inst{1} \and
Thomas Martinetz\inst{1}}
\authorrunning{K. Anderson and T. Martinetz}
%
\institute{Institute for Neuro- and Bioinformatics, University of Lübeck, Germany\\
\email{\{anderson,martinetz\}@uni-luebeck.de}}
\maketitle              
\begin{abstract}
  The ability to remove unwanted dependencies from data is crucial in various domains, including fairness, robust learning, and privacy protection. In this work, we propose an encoder-based approach that learns a representation independent of a sensitive variable but otherwise preserving essential data characteristics. Unlike existing methods that rely on decorrelation or adversarial learning, our approach explicitly estimates and modifies the data distribution to neutralize statistical dependencies. To achieve this, we combine a specialized variational autoencoder with a novel loss function driven by non-parametric nearest-neighbor density estimation, enabling direct optimization of independence. We evaluate our approach on multiple datasets, demonstrating that it can outperform existing unsupervised techniques and even rival supervised methods in balancing information removal and utility.
\end{abstract}
\section{Introduction}
Hidden statistical dependencies exist in almost every dataset. They can encode useful biases, but may also impede the learning process, like an object frequently, but not always, appearing against a specific background or a measurement systematically skewed by the equipment used to generate it. In some cases, these dependencies can even be harmful and lead to unfair or discriminatory outcomes.


In this work, we train an encoder to generate representations that are independent of a sensitive variable while preserving the original input as much as possible. Our approach extends beyond simple decorrelation, with the objective of completely neutralizing statistical dependencies. Previous methods mainly rely on adversarial strategies or loose lower bounds, both of which can be prone to miscalculations. Instead, we propose a solution for the challenging task of directly measuring and manipulating densities in continuous data, by translating the concept of a non-parametric density estimator into a differentiable loss function. The accuracy of the proposed estimator is influenced by specific distributional properties of its input. To ensure these properties are met advantageously, we combine the encoder with a specialized variational autoencoder (VAE), which guides the representation toward a suitable distribution.

\section{Related Work}
The idea of removing information from a dataset has been widely explored across various applications and has been an active area of research since 1999 \cite{tishby99IB}.

The classical information bottleneck (IB), a basis of many strategies in the field of information removal, aims to create a "compressed summary" of the data by neutralizing \textit{any} information that is redundant or irrelevant for a specified task. Variational autoencoders (VAE), whose regularization term effectively minimizes the dependence between an intermediate latent and the output, serve as an appropriate foundation \cite{tishby99IB,alemi2017IB,achille2018IB,fischer2020IB}.

Originally introduced as a \textit{privacy funnel} in \cite{makhdoumi2014privacyFunnel}, a common reformulation of the original IB minimization goal is to define a sensitive variable $S$ and only reduce information about $S$, while retaining any information not connected to it. Despite diverging from the original IB objective, many studies on targeted information removal continue to label their methods as information bottleneck and make use of VAEs in a similar fashion \cite{moyer2018variationalInvariance,poole2019variationalInvariance,rodriguez2021variationalInvariance}. While the concept can work very well, the VAE regularization loss, which in its pure form aims to remove \textit{all} dependency between input and latent, needs to be balanced carefully with the reconstruction goal.

An alternative not based on VAEs and with a less rigorous regularization goal is to incorporate contrastive learning. The strategy comes down to increasing the distance between samples with the same sensitive label, while decreasing the distance between those with different labels \cite{gupta2021contrastiveInvariance,akash2021contrastiveInvariance}. Another relaxation of the strict VAE regularization term uses subtraction to minimize the average distance between the distribution parameters returned for different sensitive labels \cite{liu2022variationalInvariance}.

In a different line, adversarial learning has emerged as a widely adopted approach in this context \cite{roy2019adversary,wang2019adversary,zhao2020adversary,madras2018adversary,creager2019adversary,jaiswal2018adversary,jaiswal2020adversary,kairouz2022adversary,cui2023adversary}. The underlying concept is straightforward: train an encoder to generate a representation while simultaneously training an adversary to detect the sensitive information from the representation. Following the standard adversarial framework, the encoder aims to maximize the adversary’s loss, thereby ensuring that the sensitive variable remains unidentifiable. 
While the adversarial approach circumvents the complexities of statistical learning, it is inherently less reliable. The resulting encoding is designed to evade detection by a specific adversary, but this does not guarantee the true removal of sensitive information. A more powerful or better-suited adversary may still recover the sensitive attribute.

Unlike general information removal approaches, some fair learning methods assume that both the sensitive attribute and the target label are known. This is particularly relevant when the target label is correlated with the sensitive attribute or when the learning process tends to overfit to unwanted biases. Some methods focus on creating embeddings that retain \textit{only} target label information while discarding sensitive attributes \cite{lu2023variationalInvariance,oh2022contrastiveInvariance}, others modify the learning process itself to mitigate bias during training, ensuring that the model does not rely on sensitive information \cite{roh2020fairLearning,kim2019fairLearning}.

To the best of our knowledge, explicit differentiable density estimation (beyond the lower bound of VAEs) has previously not been employed to neutralize dependencies. However, previous work has implemented differentiable kernel based density estimators such as maximum mean discrepancy for different tasks, such as the generation of new samples from a target distribution \cite{li2015mmdgenerative,sutherland2015mmdgenerative}. Depending on the problem, Kernel methods might yield better estimations than the nearest neighbor strategy used in this paper, but usually come with the drawback of a substantially higher computational cost with standard kernel functions, and a strong sensitivity to a bandwidth parameter. 

We found the optimization of the nearest-neighbor density estimate to be empirically more stable than kernel methods, while reliably remaining proportional to the true density (despite these favorable properties this method is nearly absent in gradient-based learning applications \cite{sablayrolles2019knn}). Exploring how classical kernel density estimation might refine our method offers a promising direction for future research.

\section{Setting}
\label{sec:setting}

This work is based on what we consider the most common scenario for information removal. The input samples $X$ are continuous and potentially unstructured, each associated with a known, discrete (or discretizable)\footnote{The current strategy focuses on binary sensitive labels, the straightforward expansion to arbitrary categorical labels is left for future work.} sensitive label $S$. Akin to most information removal strategies, our goal is to learn a representation $Z$ that minimizes the dependency on $S$ while preserving as much information about the data as possible:
\begin{align}
\min_{Z} (I(Z;S) - \beta I(Z;X)). \label{eq:setting}
\end{align}
In many cases, including the approach taken here, maximizing mutual information $I(Z;X)$ is replaced by minimizing a reconstruction distance $-d(Z, X)$. 

We chose the generic utility term $I(Z;X)$ because it enables the preprocessing of data for both supervised and unsupervised learning tasks, even when the final objective is unknown. 
An alternative formulation replaces the reconstruction loss with $I(Z;Y)$, optimizing the representation specifically for a supervised task with a known target label $Y$. While this variation is not the focus of our work, we compare our results against approaches that leverage the additional task-specific information.

\section{Strategy}
\label{sec:pipeline}
\begin{figure*}
    \begin{center}
        \input{Tikz/pipeline}
    \end{center}
    \caption{The pipeline used to translate input samples $x$ into invariant versions $x'$.}
    \label{tikz:pipeline}
\end{figure*}
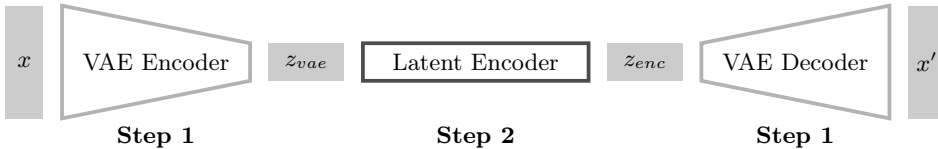

The transformation pipeline from input sample $x$ to representation $z$ consists of two stages, as illustrated in \cref{tikz:pipeline}. In a first step, a variational autoencoder (VAE) is trained to map the input into a smoothly distributed latent space (\cref{sec:vae}). Once this step is completed, both the encoder as well as the decoder of the VAE are frozen. In step two, an additional latent encoder is placed between the pretrained encoder and decoder, which now shall transform the VAE latent $z_{vae}$ into $z_{enc}$ and remove the information related to the sensitive attribute $s$. This encoder is implemented as a multi-layer perceptron (MLP) and is trained with a specialized loss function, described in \cref{sec:neighbor_density}. The resulting representation $z_{enc}$ retains the same dimensionality as $z_{vae}$ and can be passed through the pretrained VAE decoder to be reconstructed in the input space.

\subsection{Variational Autoencoders that Disentangle one Specific Variable}
\label{sec:vae}

Variational autoencoders (VAEs) are widely used for generating latent representations with a specific distribution. In this work, we employ a VAE to create the initial latent space, which is later refined using the specialized loss described in \cref{sec:neighbor_density}. To enhance their suitability for subsequent fine-tuning, we introduce a minor modification to the VAE training objective, ensuring a more robust foundation for the adaptation process.

In a standard VAE, the encoder — often referred to as \textit{probabilistic encoder} $q(z|x)$ - maps an input $x$ to a distribution over latent variables $z$. The decoder $p(x|z)$ then tries to reconstruct the original input from a latent sampled from $q(z|x)$. To enable gradient-based optimization, the reparameterization trick is applied when sampling from $q(z|x)$. Both the encoder and decoder are jointly trained to minimize
\begin{align}
\mathbb{E}_z (\underbrace{\log p(x|z)}_\text{reconstruction} - \underbrace{\mathrm{KL}(q(z|x)|p(z))}_\text{regularization}).\label{eq:vae_base}
\end{align}

Equation \ref{eq:vae_base} is derived as a lower bound for $p(x)$ (also called ELBO), but can also be understood as a combination of two opposing terms: the reconstruction term is maximal if the autoencoder successfully models the input distribution, the regularization term encourages the latent distribution to remain close to a predefined prior $p(z)$, promoting a structured and smooth latent space.

Usually, the normal distribution $\mathcal{N}(0, I)$ is taken as the prior $p(z)$, but $p(z)$ could theoretically be any distribution. To achieve specialized disentanglement, we set $p(z)$ to $\mathcal{N}(\mu, I), \mu = [s, 0, ..., 0]$. The target mean $\mu$ is set to zero for all latent dimensions except one, referred to as $z_0$. In this specific dimension, the mean is aligned with the sensitive label $s$ (therefore, technically speaking, we use a different prior for every possible value of $s$, in our case of a binary label $\mu=+1$ or $\mu=-1$). Consequently, the encoder is explicitly trained to include information about $s$ into $z_0$.
 
Using this specialized VAE introduces several key advantages for the subsequent density estimation. First, the latent representations are approximately Gaussian-distributed, ensuring a smooth space where nearby latents exhibit similar probabilities. Second, the diagonal covariance matrix of the prior encourages decorrelation and disentanglement across latent dimensions. Finally, the sensitive information is primarily compressed into $z_0$, masking $z_0$ should now hide most of the sensitive information. 
As a result, only minor refinements are needed in the next step. The task of the method introduced in the following section is to eliminate these subtle dependencies.

\subsection{Using Point Distances to Estimate Densities}
\label{sec:neighbor_density}

To fine-tune the VAE latents, we first reformulate mutual information estimation into a problem that requires only the density computation of samples across different sets. We then incorporate a density estimate based on neighbor distances into this reformulation, ultimately deriving the loss function that is to be minimized by the encoder. Note, that despite their similar names, the introduced nearest-neighbor density estimator must not be confused with the the k-nearest-neighbor classifier. The former estimates densities based on nearby neighbors, and the latter uses a majority vote of nearby target labels to partition a fixed space for classification. 

\subsubsection{KL Divergence for the Estimation of Dependency}
\label{sec:neighbor_density_KL}

In our setting, we simplify the formulation of mutual information between the representations $Z$ and the sensitive information $S$ to:
\begin{align}
    I(X;S)  
            &= \sum_{z\in Z, s\in S}p(z|s) p(s)\log\frac{p(z|s)}{p(z)}\\
            &\stackrel{*}= \sum_{z\in Z}p(z|s_z) p(s_z)\log\frac{p(z|s_z)}{p(z)}
\end{align}
Where the second equality holds because every sample $z$ in the dataset corresponds to exactly one $s := s_z$, the term $p(z|s)$ is only nonzero for $s=s_z$. 

Our datasets have been balanced to include the same number of samples for every sensitive label, $p(s) = \frac{1}{N_S}$ with $N_S$ as the total number of sensitive labels, meaning that $p(s)$ can be factored out and, for the purpose of optimization, discarded. The remaining optimization goal can therefore be written as

\begin{align}
    \text{I}(Z;S)   &\propto \sum_{z\in Z}p(z|s_z)\log\frac{p(z|s_z)}{p(z)}\label{eq:MutualInf} \\
    &= \text{KL}(p(z|s_z);p(z)).
\end{align}

with $\text{KL}(p(z|s_z);p(z))$ as the Kullback-Leibler divergence between $p(z)$ and $p(z|s_z)$. In other words, if $Z$ and $S$ are independent, the likelihood of every representation $z \in Z$ is the same when estimated in (i) the subset of representations whose sensitve label is the same as the sensitive label of $z$ and (ii) the set of all representations $Z$.

\subsubsection{Neighbor Distances for the Estimation of Probability Densities}
\label{sec:neighbor_density_KNN}
The practical implementation of \cref{eq:MutualInf} builds on a non-parametric entropy estimator originally proposed in 1987 by Kozachenko and Leonenko. Over the years, this method has undergone further analysis and refinement \cite{lombardi2016knnDensity,berrett2019knnDensity,zhao2022knnDensity}. Interestingly, despite its widespread use in applied physics, this approach has yet to be explored in the context of machine learning. 

At its core, the concept is based on a simple intuition: a point $z$ has a high probability if it is surrounded by many close neighbors. Unlike the original entropy estimator, our strategy only requires measures proportional to the density, allowing for minor simplifications and the omission of normalization steps.

\begin{figure}
    \begin{center}
        \input{Tikz/knn}
    \end{center}
    \caption{Depiction for the method described in \cref{sec:neighbor_density_KNN}. To estimate the density of $z$, one effectively counts the points that are in its circumference $\mathcal{B}$.}
    \label{tikz:KNN_density}
\end{figure}
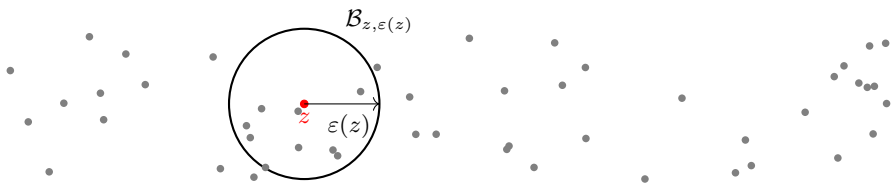

As depicted in \cref{tikz:KNN_density}, we estimate the probability density $p(z)$ for $z \in Z$, w.l.o.g. $Z \subset \mathcal{R}^2$. $\mathcal{B}_{z,\varepsilon(z)}$ denotes a circle centered on $z$ with radius $\varepsilon(z)$, total probability (or probability mass) $p(\mathcal{B}_{z,\varepsilon(z)})$ and area $\varepsilon(z)^d \cdot c$, with $c$ being the area of the unit circle and $d$ the dimensionality of the space (in the example $d=2$). The nearest-neighbor estimator is based on two assumptions:

\begin{asu}
The probability density of a neighborhood around a point is uniform.
\end{asu}
Therefore, if we know the probability mass of $\mathcal{B}_{z,\varepsilon(z)}$, we can estimate the probability density at the point $z$ that defines its center simply by 
\begin{align}\label{eq:knn1}
p(z)= p(\mathcal{B}_{z,\varepsilon(z)}) / (\varepsilon(z)^d \cdot c).
\end{align}
In other words, the probability density at a single point in $\mathcal{B}_{z,\varepsilon(z)}$ is equal to the probability mass of the area $\mathcal{B}_{z,\varepsilon(z)}$, divided by its size.

\begin{asu}
The number of samples within an area is a good estimate for the probability mass. 
\end{asu}
An area containing $M$ samples has the total probability mass $M/N$, with $N$ as the number of samples in $Z$, hence
\begin{align}\label{eq:knn2}
p(\mathcal{B}_{z,\varepsilon(z)}) = M/N.
\end{align}

Putting \cref{eq:knn1} and \cref{eq:knn2} together yields 
\begin{align}
p(z)= \frac{M/N}{\varepsilon(z)^d \cdot c}.
\label{eq:k-estimation}
\end{align}

In practice, we do not fix $\varepsilon(z)$ and count the number $M$ of neighbors within that radius. Instead, we fix $M$ and determine the smallest $\varepsilon(z, M)$ required to enclose exactly $M$ points. $\varepsilon(z, M)$ then corresponds to the distance to the $M$-th neighbor of $z$. 

\subsubsection{Combined Estimator for Mutual Information}
\label{sec:neighbor_density_loss}
We now insert our density estimator (\cref{eq:k-estimation}) into the dependency formula defined in \cref{eq:MutualInf}. As described in \cref{sec:single_dim}, we optimize each dimension of the representation separately, making $d$ in \cref{eq:k-estimation} equal to one. To simplify the formula, we omit $d$ in the following description.

For a representation $z$ with the sensitive label $s_z$, let $p(z)$ be the probability density estimated at $z$ in the set of all representations, and $q(z)=p(z|s_z)$ be the estimate for the probability density at $z$ in the subset of representations that all correspond to label $s_z$. $N_p$, and $N_q$ refer to the respective set sizes, $\varepsilon_p(z, M)$ and $\varepsilon_q(z, M)$ to the distance to the $M$-th neighbor of $z$ in $p$ or $q$. Since both measurements are in the same sample space and using the same metric, the area of a unit-sphere $c$ is identical, and we obtain 
\begin{align}
    \frac{p(z|s_z)}{p(z)}   &\approx \left( \frac{M/N_q}{\varepsilon_q(z, M) \cdot c}\right)/\left( \frac{M/N_p}{\varepsilon_p(z, M) \cdot c}\right)
                             = \frac{N_p \cdot \varepsilon_p(z, M)}{N_q \cdot \varepsilon_q(z, M)}.
\end{align}
    
A variation that is more stable is to adjust $M$ to accommodate the different set sizes $N$. Intuitively, if there are twice as many samples in $p$, the number $M$ of neighbors in the same area should be twice as high to represent the same probability. This leads to
\begin{align}\label{eq:p_div_q}
    \frac{p(z|s_z)}{p(z)} \approx \frac{\varepsilon_p\left(z, \frac{N_p}{N_q}M\right)}{\varepsilon_q\left(z, M\right)}.
\end{align}
With \eqref{eq:MutualInf} we obtain as an estimate for the mutual information 
\begin{align}
    \text{I}(Z;S)   \propto \sum_{z\in Z}p(z|s_z)\log\frac{p(z|s_z)}{p(z)}
                    \approx \sum_{z\in Z}p(z|s_z)\log\frac{\varepsilon_p\left(z, \frac{N_p}{N_q}M\right)}{\varepsilon_q\left(z, M\right)}.
                     \label{eq:final_loss}
\end{align}
The mutual information as the final loss we want to minimize can now be determined by Monte-Carlo sampling.

\section{Implementation Details}
The loss function formulated in  \cref{eq:final_loss} closely aligns with the objective of independence. A value of zero indicates that the representation $Z$ is independent of $S$, with a comparatively\footnote{Compared to, e.g. the trade-of of pure VAE solutions.} low estimation error. 
However, the exact loss formulation presents a new challenge: finding a global minimum is non-trivial. To enhance training stability and convergence, we incorporate several implementation optimizations.

\subsection{Optimizing each Dimension Separately}
\label{sec:single_dim}
The VAE pretraining disentangles the dimensions in the latent space, and by using appropriate priors the sensitive information is shifted into $z_0$. However, without explicit constraints, the MLP used for fine-tuning with its fully connected layers naturally tends to re-entangle the input dimensions. We observed that when the latent encoder is trained on the entire VAE latent space, it can inadvertently redistribute sensitive information from $z_0$ back into the remaining latent dimensions, counteracting the intended disentanglement. 

Experiments revealed that the performance of the pipeline can be improved by instead training a separate latent encoder for each individual latent dimension. This alteration offers another key advantage: the now one-dimensional input reduces estimation errors arising from the assumptions outlined in \cref{sec:neighbor_density_KNN}.

The sum over the mutual information estimated for every dimension serves as an upper bound to the mutual information between the entire latent $z$ and the sensitive label $S$, if each each dimension of the latent is treated as independent:
\begin{align*}
    \sum_i I(z_i;S) = \sum_i H(z_i) - \sum_i H(z_i|S) \stackrel{*}{\geq} H(z) - H(z|S) = I(Z;S).
\end{align*}
While the requirement for this bound can not be guaranteed, it coincides with the optimization goal of the VAE regularization term. 

\subsection{Reducing the Effect of Noisy Samples}
For smaller values of $M$, a few noisy samples can disproportionately inflate the loss in \eqref{eq:final_loss}. To mitigate this effect, we apply Gaussian kernel smoothing to the distance computation. Instead of using only the distance to the $M$-th neighbor, we compute a weighted combination of the distances to the  $[M-2, M-1, M, M+1, M+2]$ nearest neighbors. Similarly, density estimation is stabilized by averaging over multiple values of $M$, further reducing sensitivity to noise.

\subsection{Pretraining with Squared Distances}
Optimizing the fraction in \cref{eq:final_loss} for two sets with large divergences can lead to loss explosions, preventing effective encoder training. To address this issue, we substitute
\begin{align}
\log \frac{p(z|s_z)}{p(z)} \quad\text{by}\quad \left(1 - \frac{p(z|s_z)}{p(z)}\right)^2. 
\end{align}
In both cases, the optimal solution is achieved when $p(z|s)=p(z)$. However, the squared distance formulation offers greater numerical stability and allows for straightforward gradient computation. In the beginning of the training phase, $p(z|s)$ can be very small, making it useful to even fall back to the squared distance ($(p(z|s) - p(z))^2$ instead of $(1 - \frac{p(z|s)}{p(z)})^2$). Again, both have the same optimum. We combine the two versions by training for $(p(z|s) - p(z))^2$ first and optimizing $(1 - \frac{p(z|s)}{p(z)})^2$ once the remaining divergence is small enough.

\section{Experiments}
The evaluation is performed on different image datasets, and relies on dataset-specific target labels to estimate the utility of the transformed data. Note that for the \textit{unsupervised} setting, those target labels are not known while training the encoder, creating a representation with a more general utility.  

We compare our results against the top performing methods (with reproducible results) for each of the different categories: VAEs 
\cite{lu2023variationalInvariance,liu2022variationalInvariance} 
adversarial approaches \cite{roy2019adversary} 
and contrastive loss terms \cite{akash2021contrastiveInvariance}. 
Some methods can only be applied to a \textit{supervised} setting: the target labels that we use to estimate the utility have to be known while training the encoder, to create a representation specific to a given task.

Consistent with related work, we use the performance of an \textit{attacker} (implemented as an MLP), trained to extract the sensitive information from the transformed dataset. Our implementation is available under \url{https://github.com/ka-anderson/VAE-NN-debias}.

\subsection{Datasets}
As a simple toy dataset, we add gray background shapes to MNIST digits \cite{dsMnist}, to simulate unwanted background information. The sensitive label is the information whether the background shape is a square or a circle. To evaluate the utility, we measure how accurately a classifier can identify the original digit label from the reconstructed image.

For a more challenging task, we apply our method to the FFHQ dataset \cite{Karras2019StyleGAN1FFHQ}, a dataset containing human portraits, using attribute information from \cite{FFHQattributes}. The sensitive variable is the gender of the depicted person (in this dataset given as a binary value). As a metric for the utility of the outcome, we calculate the accuracy of a classifier trained to identify the angle of the head and the information whether the depicted person is smiling or not.

Our final challenge is to remove information from chest radiographs in the CheXpert (small) dataset \cite{dsChexpert}. We aim to mask information about the presence of support devices (e.g. pacemakers). Utility classification targets are the binary observations: "Lung Opacity", "Edema" and "Pleural Effusion", positive in 49\%, 26\% and 40\% of the data, respectively. The implementation repository provides further details about our choice of target attributes.

\begin{table*}
    \caption{Accuracy of a trained classifier on transformed data. Italic text indicates sensitive variables, whose accuracy is intended to be low. To highlight the tradeoff between information removal and utility, we also report the distance between sensitive and target accuracy. The baseline is the classifier accuracy on the original data. Accuracy is measured in percent, except for pose, which is reported as the mean absolute distance between output and target rotation angles. \\}
    
    \makebox[\linewidth][c]{
    \begin{tabular}{ll|l|lll|lll}
    \hline
                           &                     & \multirow{1}{*}{Base} & \multicolumn{3}{c|}{Supervised}  & \multicolumn{3}{c}{Unsupervised}     \\
                           &                     &                           & VAE   & Contrastive & Adversary & VAE   & Contrastive & \textbf{Ours} \\ 
                           &                     &                           & Lu 2023   & Akash 2021 & Roy 2019 & Liu 2022   & Akash 2021 &  \\ \hline
    \multirow{3}{*}{MNIST} & \textit{background} & 100                       & 50.7  & 62.2         & 56.5      & 51.6  & 52.7         & 51.8          \\
                           & digit               & 99.7                      & 98.1  & 97.8         & 97.0      & 95.2  & 96.9         & 96.8          \\
                           & \text{\textbar}\textit{backgr.} - digit\text{\textbar}    & 0.3                        & 47.4  & 37.6         & 40.5      & 43.6  & 44.2         & 45            \\\hline
    \multirow{4}{*}{FFHQ}  & \textit{gender}     & 86.2                      & 56.2  & 62.7         & 65.5      & 62.6  & 65.1         & 58.2          \\
                           & smile               & 87.9                      & 84.0  & 78.2         & 86.6      & 59.6  & 59.0         & 74.8          \\
                           & pose                & 0.037                     & 0.065 & 0.055        & 0.055     & 0.056 & 0.054        & 0.051         \\
                           & \text{\textbar}\textit{gen.} - smile\text{\textbar}        & 1.7                       & 27.8  & 15.5         & 21.1      & 3     & 6.1          & 16.6          \\ \hline
   \multirow{3}{*}{CheXpert} & \textit{devices}              & 67.4                        & 54.0  & 64.3         & 58.9      & 63.8  & 62.9           & 66.1           \\
                          & observations         & 74.4                        & 66.1  & 72.6         & 66.4      & 69.9  & 70.0           & 74.0            \\
                          & \text{\textbar}\textit{dev.} - observ.\text{\textbar} & 7.0                      & 12.1  & 8.3          & 7.4       & 6.1   & 7.1           & 7.9           \\ \hline
    \end{tabular}
    }
    \label{tab:results}
    
\end{table*}
\subsection{Results}
The classification accuracies computed for the three datasets are depicted in \cref{tab:results}. On MNIST, our method outperforms not only the unsupervised alternatives, but even two out of three supervised methods. 

Clearly, the larger portrait images of FFHQ are more difficult to transform. Yet once again, our method achieves a better tradeoff than the two unsupervised models, and even surpasses the tradeoff of the supervised contrastive model. The overall lower performance of unsupervised methods might be explained with correlations between the \textit{gender} and \textit{smile} - when not specifically aiming to preserve the given target information, the removal of gender information can inadvertently change the expression of a depicted person as well. Additionally, the target task itself is more difficult: the MLP classifier can not achieve a perfect accuracy even when trained on the unchanged images.

While the MLP classifier performance on the most challenging dataset, CheXpert, again confirms our method as the strongest unsupervised approach, it also highlights current limitations of all information removal methods on complex datasets. If the information is not easily discernible in the untransformed data, alterations to the dataset are also more challenging. 

The reported accuracies are the maximum test accuracies observed within ten epochs. When comparing not the best, but the final accuracy (after exactly ten epochs without early stopping), our method outperforms similar work by an even larger margin, indicating a model with reduced overfitting tendency. Exemplary results for the CheXpert dataset are presented in \cref{tab:full_training_results}.

\begin{table*}
    \caption{Accuracy of a trained classifier on transformed CheXpert data. Unlike \cref{tab:results}, not the highest, but the last accuracy after 10 epochs is reported.\\}
    \centering
    \begin{tabular}{l|l|lll|lll}
    \hline
                                                & \multirow{1}{*}{Base} & \multicolumn{3}{c|}{Supervised}  & \multicolumn{3}{c}{Unsupervised}     \\
                                                &                           & VAE   & Contrastive & Adversary & VAE   & Contrastive & \textbf{Ours} \\ 
                                               &                           & Lu 2023   & Akash 2021 & Roy 2019 & Liu 2022   & Akash 2021 &  \\ \hline
    \textit{devices}              & 66.5                          & 54.0    & 59.4        & 54.9                           & 60.7     & 62.9        & 60.3          \\
                           observations        & 70.5                          & 66.1    & 63.7        & 66.1                           & 69.8     & 67.8        & 72.6          \\
                           \text{\textbar}\textit{dev.} - observ.\text{\textbar} & 4.0                           & 12.1    & 4.3         & 11.2                           & 9.1      & 4.8         & 12.4           \\ \hline
    \end{tabular}
    \label{tab:full_training_results}
\end{table*}

\subsection{Robustness against Noisy Labels}
While the main focus of our work is the removal of information for the purpose of fairness and privacy, the fulfillment of this task can also come with the added benefit of improving model generalization: especially if the target labels are noisy, a trained network might overfit to spurious image features rather than the intended target. Selectively removing irrelevant information from a dataset reduces this risk. 

To illustrate this effect, we train an MLP to detect the digit label from the previously used MNIST images with random backgrounds. Noisy labels are simulated by replacing a percentage of the training labels with random values. The validation labels remain unchanged.

\begin{figure}
    \centering
    \includegraphics[width=0.5\linewidth]{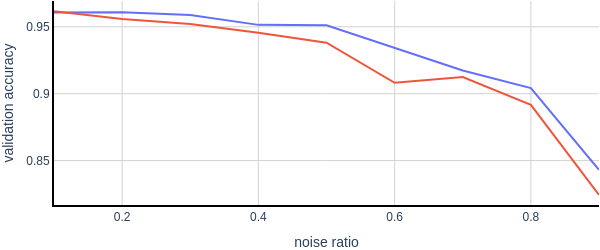}
    \caption{Validation accuracy of an MLP on MNIST with noisy labels. The blue (upper) line shows performance after background removal, while the orange line is the baseline. A noise ratio of 0.2 means 20\% of training labels were randomly replaced.}
    \label{fig:noise_acc}
\end{figure}

As seen in \cref{fig:noise_acc}, removing background information yields higher classification accuracies, even though the target digit labels have never been incorporated into the information removal pipeline.

\begin{figure}
    \centering
    \includegraphics[width=.8\linewidth]{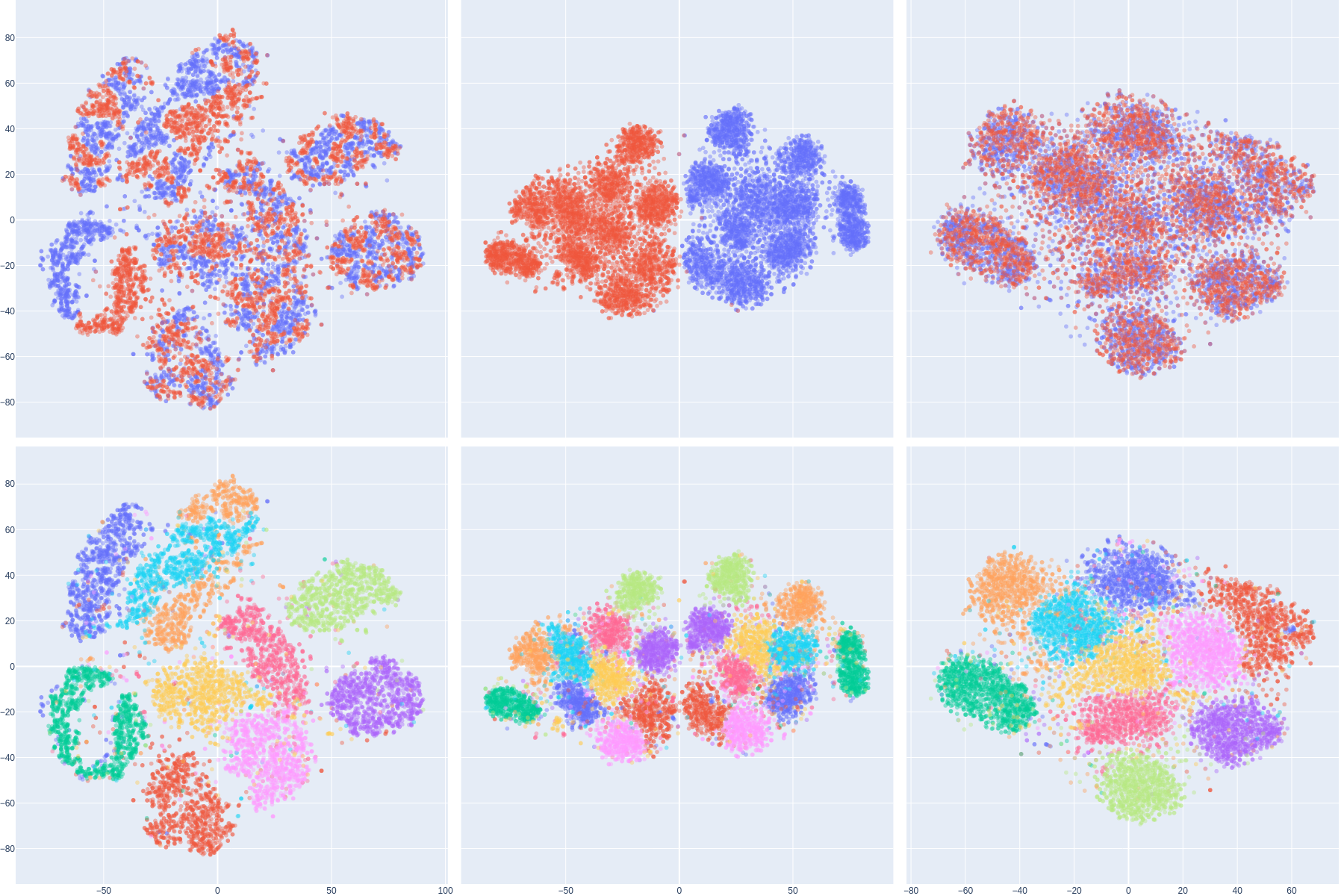}
    \caption{T-SNE embedding for the MNIST dataset with backgrounds. The top row is colored by background label, bottom by digit. From left to right, the columns depict the original data, the VAE latent ($z_{vae}$) and the encoded latent ($z_{enc}$).}
    \label{fig:mnist_tsne}
\end{figure}
\subsection{T-SNE Visualization}
The effect of our modified VAE and the density-based loss can be visualized with a two-dimensional t-SNE embedding \cite{tsne}. Figure \ref{fig:mnist_tsne} depicts the test samples of the MNIST dataset with random backgrounds. For the input images ($x$ in \cref{tikz:pipeline}) in the first column, there is a clear distinction between the two background shapes. The second column shows an embedding of the latents computed by the our VAE, $z_{vae}$. Since the sensitive part of the latent ($z_0$) has not yet been masked, the two background options divide the data into two distinct regions. Interestingly, the arrangement of digit labels is mirrored. Finally, when masking the sensitive $z_0$ and encoding the latent with the additional encoder, the two background shapes are mapped to the same areas of the embedding space. 

Throughout the encoding process, the digit labels remain well separated, while the sensitive information is shuffled.

\subsection{Ablation Study}
\label{sec:ablation}
\begin{figure}
    \centering
    \includegraphics[width=.9\linewidth]{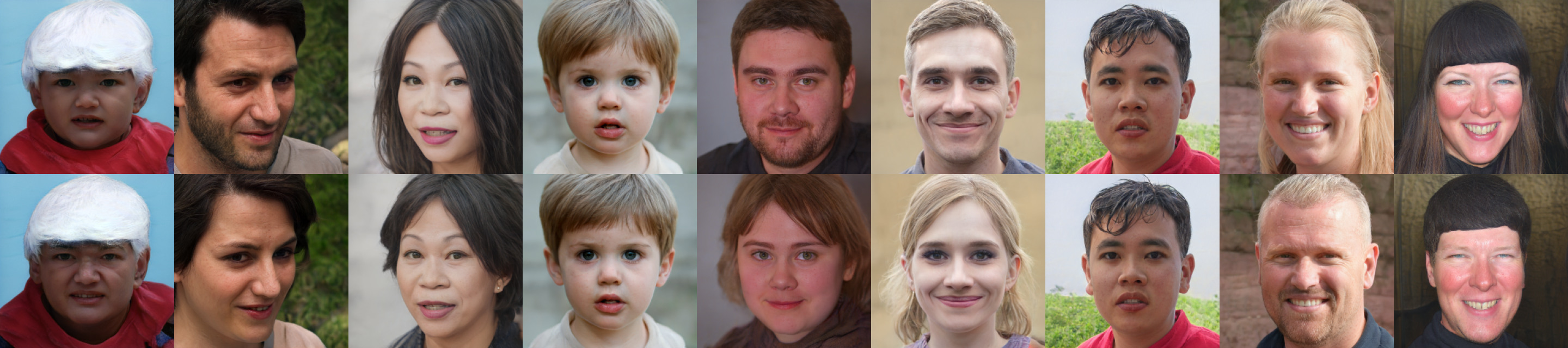}
    \caption{Images reconstructed from StyleGAN latents (see \cref{sec:ablation}). The bottom row shows images reconstructed from latents that have been translated using our nearest-neighbor divergence, to remove gender information. Note how the gender is not "removed", but rather shifted randomly for every image.}
    \label{fig:nn_without_vae}
\end{figure}

\begin{table}[]
\caption{Highest test accuracies of networks trained to predict from images transformed by the specialized VAE, but not the nearest neighbor latent encoder.\\}
\centering
\begin{tabular}{l|lll} \hline
      & MNIST & FFHQ & CheXpert \\ \hline
\textit{sensitive} & 55.1 & 80.4 & 67.1\\
target  & 97.0 & 85.4 & 74.0 \\ 
\text{\textbar}\textit{sensitive} - target.\text{\textbar}  & 41.9 & 5.0 & 6.9 \\  \hline
\end{tabular}
\label{tab:vae_without_nn}
\end{table}
Both the VAE and the nearest neighbor divergence are important to achieve the results presented in the previous section. The accuracies for latents $z_{vae}$ computed by the VAE after masking the first dimension but without employing the additional latent encoder, are depicted in \cref{tab:vae_without_nn}. The tradeoff (as the distance between sensitive and target accuracy) is reduced for all three datasets - the added latent encoder contributes an average improvement of 5.2 percentage points in the tradeoff distance.

To evaluate the importance of the VAE, we replace the VAE by StyleGAN \cite{Karras2020StyleGAN2} combined with \textit{Hyperstyle} \cite{alaluf2022hyperstyle}, with \textit{Hyperstyle} for encoding and StyleGAN for decoding the latent. Both are pretrained. The StyleGAN latents are transformed by our MLP to optimize the nearest neighbor divergence loss. The outcome of the strategy is depicted in \cref{fig:nn_without_vae}. 

Perceptually, the adjusted method results in impressively realistic images, partially thanks to the sophisticated StyleGAN latent space. However, a network trained as an attacker to extract gender information from the translated images still achieves an accuracy of 76\%. Increasing the relative weighting of the information removal loss term (i.e. lowering $\beta$ in \cref{eq:setting}) decreases the visual quality of the outcome, but not the attacker accuracy. As indicated in \cref{sec:neighbor_density}, the nearest neighbor loss term strongly depends on relatively smooth distributions (if two latents are spatially close, their probability is assumed to be similar). It is our assumption that this requirement is not fulfilled in all areas of the reconstructed StyleGAN latent space, given that this target is not part of the GAN formulation.

\section{Conclusion}
Training for true statistical independence is inherently challenging, primarily due to the difficulty of estimating probability distributions in continuous spaces. We address this issue by combining a specially trained variational autoencoder with a novel mutual information approximation method. This approach enables the transformation of arbitrary data into a latent representation with masked sensitive information, optionally followed by the reconstruction of the transformed latent back into the original data space.

Our results demonstrate that this method achieves a superior trade-off between information removal and utility compared to previous state-of-the-art unsupervised approaches. Notably, it performs on par with supervised methods, despite not relying on target label information.

Since the VAE decoder can map the transformed latent representation back into the original space, our method enables training a model on the bias-free transformed dataset while still allowing deployment on real (untransformed) data. By preventing the model from learning and internalizing unwanted biases during training, our approach ensures that its predictions on real data remain unaffected by spurious correlations, leading to fairer and more robust decision-making.


%
%
%
\bibliographystyle{splncs04}
\bibliography{bibliography}
\end{document}

%% file: Tikz/pipeline.tex
\tikzset{arrow/.style={-stealth, thick, draw=gray!80!black}}

\begin{tikzpicture}

    \node[fill=black!20, minimum width=.5cm, minimum height=1.5cm] (X) at (.25, .5) {$x$};

    \draw[black!30, line width=.5mm] (.75,-.25) -- ++(2.5,.5) -- ++(0,0.5) -- ++(-2.5,.5) -- cycle;
    \node[align=center] at (2,.5) {VAE Encoder};
    \node[align=center] at (2,-.5) {\textbf{Step 1}};

    \node[fill=black!20, minimum width=1cm, minimum height=.5cm] (rbase) at (4, .5) {$z_{vae}$};
    
    \node[align=center, draw=black!70, line width=.5mm, minimum width=3cm, minimum height=.5cm] (lenc) at (6.25, .5) {Latent Encoder};
    \node[align=center] at (6.25,-.5) {\textbf{Step 2}};
    
    \node[fill=black!20, minimum width=1cm, minimum height=.5cm] (rtrans) at (8.5, .5) {$z_{enc}$};
    
    \draw[black!30, line width=.5mm] (9.25,.25) -- ++(2.5,-.5) -- ++(0,1.5) -- ++(-2.5,-.5) -- cycle;
    \node[align=center] at (10.5,.5) {VAE Decoder};
    \node[align=center] at (10.5,-.5) {\textbf{Step 1}};

    \node[fill=black!20, minimum width=.5cm, minimum height=1.5cm] (X2) at (12.25, .5) {$x'$};


\end{tikzpicture}

%% file: Tikz/knn.tex
\usetikzlibrary {arrows.meta} 
 
\begin{tikzpicture}
    
    \draw[thick](4,1) circle (1);
    \node[align=left] at (5,2.1) {$\mathcal{B}_{z,\varepsilon(z)}$};
    
    \filldraw[red] (4,1) circle (.05) node[anchor=north]{$z$};
    
    \draw[->] (4,1) -- (5,1);
    \node[align=center] at (4.6,.7) {$\varepsilon(z)$};

    \pgfmathsetseed{1}
    \foreach \p in {1,...,50}{
        \fill[gray] (6*rand+6,1*rand+1) circle (0.05);
    }

\end{tikzpicture}